\pdfoutput=1

\documentclass[11pt]{article}

\usepackage[final]{acl}
\usepackage{float}  
\usepackage{times}
\usepackage{latexsym}
\usepackage{hyperref}
\usepackage{url}
\usepackage{authblk}
\usepackage[T1]{fontenc}

\usepackage[utf8]{inputenc}

\usepackage{microtype}

\usepackage{inconsolata}

\usepackage{graphicx}

\usepackage{booktabs} 
\usepackage{adjustbox} 
\usepackage{multicol} 
\usepackage{lscape} 
\title{MoDEM: Mixture of Domain Expert Models}

\author{\textbf{Toby Simonds}\hspace{1em}\textbf{Kemal Kurniawan}\hspace{1em}\textbf{Jey Han Lau}
\\
The University of Melbourne
\\
\texttt{tsimonds@student.unimelb.edu.au}
\\
\texttt{\{kurniawan.k, laujh\}@unimelb.edu.au}
}

\begin{document}
\maketitle
\begin{abstract}
We propose a novel approach to enhancing the performance and efficiency of large language models (LLMs) by combining domain prompt routing with domain-specialized models. We introduce a system that utilizes a BERT-based router to direct incoming prompts to the most appropriate domain expert model. These expert models are specifically tuned for domains such as health, mathematics and science. Our research demonstrates that this approach can significantly outperform general-purpose models of comparable size, leading to a superior performance-to-cost ratio across various benchmarks. The implications of this study suggest a potential paradigm shift in LLM development and deployment. Rather than focusing solely on creating increasingly large, general-purpose models, the future of AI may lie in developing ecosystems of smaller, highly specialized models coupled with sophisticated routing systems. This approach could lead to more efficient resource utilization, reduced computational costs, and superior overall performance.
\end{abstract}

\section{Introduction}



Domain-specific models have demonstrated encouraging performance across various fields, often surpassing state-of-the-art general models in their respective domains. In mathematics, models like Qwen 2 72B Math \citep{yang_qwen2_2024} and DeepSeek Math \citep{shao_deepseekmath_2024} have shown superior performance, while in code generation, specialized models such as Code Llama and CodeMistral exhibit significant improvements over comparable general-purpose models~\citep{ai_codestral_2024}. Also, \citet{zhao_lora_2024} found that models with fewer than 8 billion parameters, when fine-tuned for specific tasks, can rival or even outperform larger models like GPT-4 in certain domains.

\begin{figure}
    \centering
    \includegraphics[width=1\linewidth]{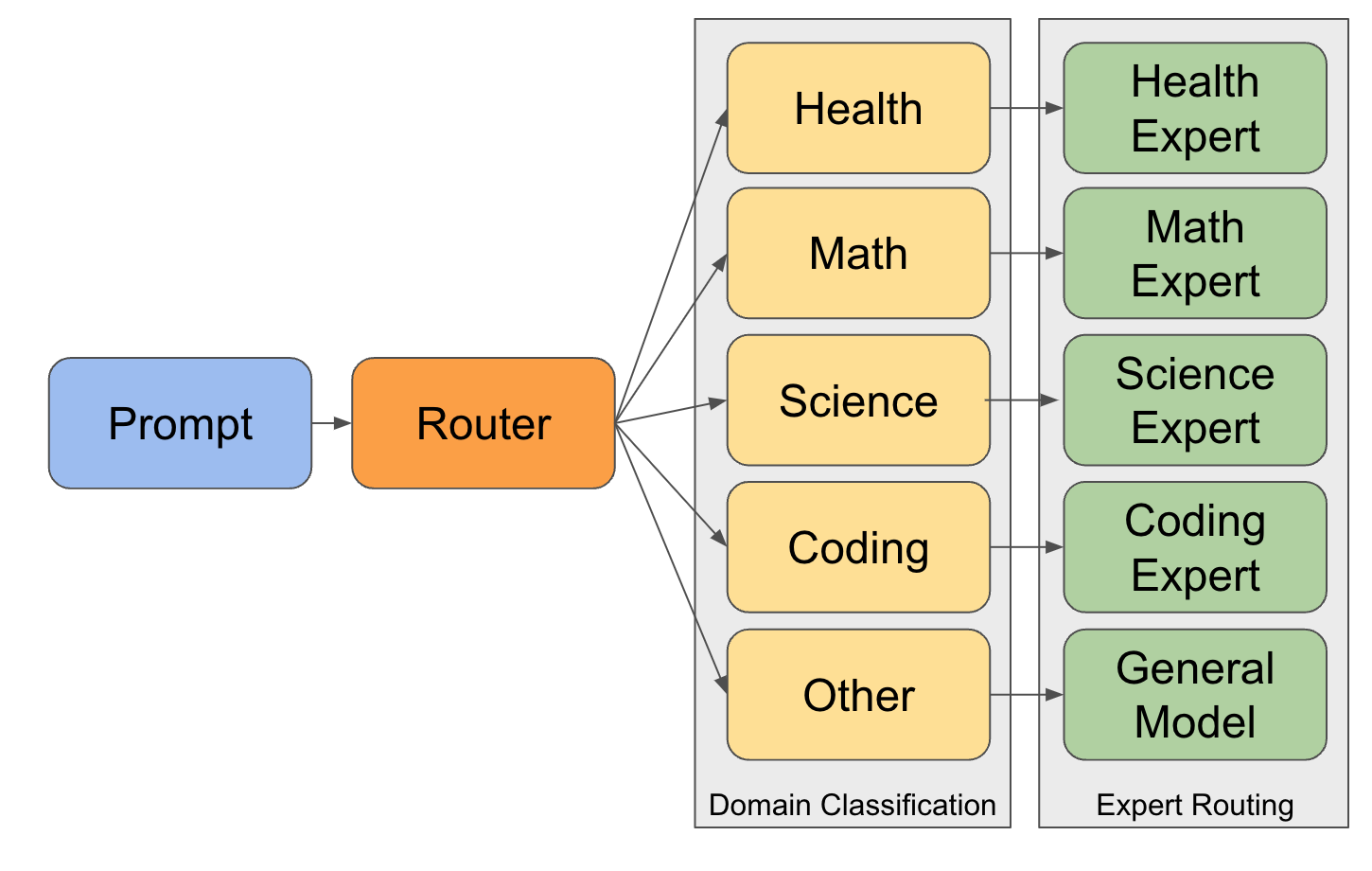}
    \caption{MoDEM architecture diagram}
    \label{fig:enter-label}
\end{figure}

Despite the promise of domain-specific AI models, a significant gap exists in integrating these specialized models into a comprehensive and versatile framework. The AI community faces a crucial challenge: how to harness the power of domain-specific models across diverse tasks without sacrificing the versatility of general-purpose models.

 We propose MoDEM (Mixture of Domain Expert Models) to address this. At its core, MoDEM consists of two main components: a router and a collection of domain-specific expert models (Figure \ref{fig:enter-label}). The router is designed to classify incoming prompts or queries, determining which domain they best fit into. Once classified, the prompt is then directed to the expert model specialized in that particular domain. This approach allows us to harness the superior performance of domain-specific models while maintaining the ability to handle a wide range of tasks.
By leveraging smaller specialized models, we achieve state-of-the-art results in various domains without the computational overhead of larger general-purpose models. This approach dramatically lowers inference costs, as only the relevant expert model is activated for each query. The result is a highly efficient system that delivers strong performance while minimizing resource utilization.

MoDEM key advantage lies in its ability to train and integrate models separately, offering significant benefits in development efficiency and system capabilities. This approach allows for independent optimization of domain experts, facilitates parallel development, and enables easy integration of new models.
The modular design ultimately allows for customization across various industries and applications.

To summarise, our main contributions are:
\begin{itemize}
    \item We propose an architecture for creating a lightweight router system that effectively directs prompts to domain-specific expert models.
    \item We demonstrate that domain-based routing to specialized experts can produce state-of-the-art results with significant inference cost reduction.
\end{itemize}

\section{Related Work}
Mixture of Experts (MoE) is a machine learning technique that combines multiple specialized models or "experts" to solve complex tasks. In the context of language models, MoE approaches have been explored to enhance both performance and efficiency. There are primarily two categories of MoE implementations in current research:

\subsection{Integrated MoE Architectures} Sparse Mixture of Experts (MoE) transformers is first introduced by \citet{shazeer_outrageously_2017} and further developed in models such as GShard \citep{lepikhin_gshard_2020} and Switch Transformers \citep{fedus_switch_2022}, which integrate expert modules within a single model architecture. These methods use a gating mechanism to dynamically route tokens or layers to different expert sub-networks during training and inference, significantly improving model efficiency by activating only a subset of experts. However, these approaches encounter challenges such as training instability, architectural complexity, and load balancing issues \citep{li_locmoe_2024}.

\subsection{Multi-Model Routing Systems} Recent research has explored systems that leverage multiple distinct language models rather than sub-networks within a single architecture. For example, HuggingGPT \citep{shen_hugginggpt_2023} breaks tasks into subtasks and routes them to different specialized models. Another approach, RouteLLM \citep{ong_routellm_2024}, aims to optimize the cost-performance trade-off by selecting between two pre-trained models for different tasks. MoDEM is different to HuggingGPT and RouteLLM in that our approach routes questions into \textit{domains} such as mathematics or health; this is a contrast to HuggingGPT where it routes based on tasks (e.g. OCR) or RouteLLM which attempts to directly predict different models performances in order to attempt to route to the best model. 

\section{Methodology}

\subsection{Benchmarks}
\label{sec:benchmarks}

We use the following evaluation benchmarks to measure the performance of MoDEM: MMLU, MMLU Pro, HumanEval, College Math, Math, GSM8k, and Olympiad Bench. These benchmarks were chosen to provide a balanced distribution of domain-specific and general tasks, ensuring a comprehensive evaluation across diverse areas of expertise.

\textbf{MMLU} \cite{hendrycks_measuring_2021} (Massive Multitask Language Understanding) is a general-purpose benchmark designed to test a model’s proficiency across 57 subjects, including STEM, humanities, social sciences, and more. The questions are in multiple-choice format, covering a broad range of domains to evaluate the model's versatility.

\textbf{MMLU Pro} \cite{wang_mmlu-pro_2024} is an extension of MMLU that focuses on more advanced topics and professional-level knowledgeg. It uses multiple-choice questions similar to MMLU, but with more specialized and higher-level content.

\textbf{GPQA} \cite{rein_gpqa_2023} is designed to evaluate models on advanced topics and professional-level knowledge across a wide array of science domains. 

\textbf{HumanEval} \cite{chen_evaluating_2021} assesses code generation capabilities by providing programming problems that the model must solve. It focuses on domain-specific knowledge within the programming domain, using open-ended coding tasks that require the model to generate functioning code.

\textbf{College Math} \cite{liu_mathbench_2024} evaluates a model’s understanding of undergraduate-level mathematics on open ended problems, covering topics such as calculus, linear algebra, and probability. 

\textbf{MATH} \cite{liu_mathbench_2024} is a more general benchmark that covers a wide range of math topics at varying levels, including elementary arithmetic, algebra, and more complex problem-solving tasks. 

\textbf{GSM8k} \cite{cobbe_training_2021} (Grade School Math 8k) is a benchmark that evaluates mathematical reasoning skills on open ended problems, specifically targeting grade-school level word problems.

\textbf{Olympiad Bench} \cite{he_olympiadbench_2024} includes challenging open ended math and science problems typically found in international Olympiad competitions.

Of these benchmarks, MMLU, MMLU Pro and GPQA rely on multiple-choice questions (MCQ) to evaluate the model’s proficiency across various domains, including general knowledge and professional-level topics. In contrast, HumanEval, College Math, Math, GSM8k, and Olympiad Bench focus on open-ended questions.




\subsection{Router}

We now describe the router, a key component used for directing incoming queries to the most appropriate domain-specific expert model.

\subsubsection{Router Architecture}
We used Microsoft DeBERTa-v3-large~\citep{he_debertav3_2023}, a 304 million parameter model, and fine-tuned it for our specific routing task. The model was fine-tuned to predict the domain of the input prompt (e.g.,\ Math).
We chose DeBERTa-v3-large due to its successful application in classification tasks.
With our largest expert models containing up to 73B parameters, the router represents only about 0.42\% of the largest expert's size. This ratio ensures that we're not spending disproportionate computational resources on routing.

\subsubsection{Domain Selection}
The domains selected for our study were the following:
Math, Health, Science, Coding and Other. Other represented domains outside of the selected domains.
These domains were chosen based on the availability of high-quality specialized models that consistently outperform general-purpose models. They also represent a diverse range of tasks and have significant real-world applications, ensuring that the routing system demonstrates versatility across various areas.

\subsubsection{Training Data}
For the router, we curated a set of diverse and comprehensive training data covering multiple domains; full list of datasets for each domain is given in Table \ref{tab:comprehensive_training_datasets}. Our focus was on selecting datasets that capture a broad range of tasks, and complexities within each domain to ensure thorough representation and variety. This approach ensures that our router is exposed to a variety of query formulations and problem types, enhancing its ability to accurately classify and route a broad range of real-world queries. We also use data from the benchmarks, specifically Math, GPQA, GSM8k and HumanEval (Section \ref{sec:benchmarks}), but only from their training partition. Note that we do not use any data from MMLU or MMLU Pro.

To ensure balanced representation across different domains, we implemented a data pruning protocol. A maximum threshold of 30,000 instances per dataset in each domain was applied to Math, Health, and Science while Other and Coding was allowed up to 100,000 entries per dataset. This decision was made because some datasets contained repetitive data, whereas the coding and other benchmarks featured more diverse and varied datasets. We down-sampled some coding datasets because they are over represented in the training set.
This methodology aimed to create a comprehensive training corpus that prevents any single source from dominating the learning process, thereby optimizing the model's ability to generalize across diverse tasks and knowledge domains. Table \ref{tab:final_data_distribution} outlines total number of training instances in each domain.

To further enhance the diversity and coverage of our dataset, we employed synthetic data generation using the Llama 3.1 405B model~\cite{dubey_Llama_2024} .
This step was crucial in addressing a significant gap we identified in existing datasets: a scarcity of casual, conversational questions that were clearly classified by domain. We found that while many datasets provided structured, formal queries, they lacked the natural language and varied scenarios typical of real-world interactions. We first created a hand-crafted dataset of 100 examples of conversation-style questions for each domain.\footnote{By ``conversation-style'', we refer to questions that simulate a more natural, interactive dialogue, as opposed to traditional fact-based or direct question-answer formats. 
} We selected a wide array of question content within each domain. We then prompted Llama 405B to generate 100 questions for each hand-crafted examples, resulting in a total of 10,000 synthetic examples for each domain.\footnote{Temperature set to 1.0 to ensure more diverse dataset \cite{noauthor_synthetic_nodate}.}
We found that incorporating hand-crafted examples into the model not only produced outputs closely aligned with our desired question format but also introduced a greater diversity of questions. When rerunning the same prompt without these hand-selected examples, the model would often generate similar outputs, lacking variety. 


Here are some examples of the handcrafted dataset:
\begin{itemize}
    \item \textbf{Math}: \textit{"I'm out with 4 friends and our total bill is \$137.50. We want to leave a 15\% tip. How much should each person pay if we split it evenly?"}
    \item \textbf{Health}: \textit{"I've had this annoying sore throat for about 4 days now. It's not super painful, but it's definitely there, especially when I swallow."}
    \item \textbf{Science}: \textit{"Can you explain how microwaves work?"}
\end{itemize}


Given the training data (data in Table \ref{tab:comprehensive_training_datasets} and the synthetic data) for each domain, we fine-tuned DeBERTa to classify the domain given an input instance. The fine-tuning was performed with a configuration of 1 epoch, a batch size of 32, and a learning rate of 1e-5. The model was trained on an A100 GPU for 1 epoch.


\begin{table}
\centering
\footnotesize
\begin{tabular}{@{}ll@{}}
\toprule
\textbf{Domain} & \textbf{Datasets} \\
\midrule
Math & 
\begin{tabular}[t]{@{}l@{}}
TIGER-Lab/MathInstruct \\
lighteval/MATH \\
allenai/math\_qa \\
openai/gsm8k \\
camel-ai/math \\
meta-math/MetaMathQA \\
deepmind/math\_dataset/algebra\_\_linear\_1d \\
deepmind/math\_dataset/algebra\_\_polynomial\_roots \\
deepmind/aqua\_rat \\
AI4Math/MathVerse
\end{tabular} \\
\addlinespace
Health & 
\begin{tabular}[t]{@{}l@{}}
nlpaueb/biomrc \\
iari/HumGen\_Clinical\_Notes \\
medmcqa \\
lavita/ChatDoctor-HealthCareMagic-100k
\end{tabular} \\
\addlinespace
Science & 
\begin{tabular}[t]{@{}l@{}}
bigbio/pubmed\_qa \\
derek-thomas/ScienceQA \\
allenai/sciq \\
bigscience/P3 \\
ai2\_arc \\
nlpaueb/biomrc \\
allenai/scitldr \\
tdiggelm/climate\_fever \\
medmcqa \\
Idavidrein/gpqa \\
allenai/scifact \\
allenai/scirepeval
\end{tabular} \\
\addlinespace
Coding & 
\begin{tabular}[t]{@{}l@{}}
codeparrot/apps \\
bigcode/the-stack \\
nuprl/MultiPL-E \\
code\_x\_glue\_ct\_code\_to\_text \\
deepmind/code\_contests \\
huggingface/codecompetitions \\
openai/openai\_humaneval \\
bigcode/humanevalpack \\
defect\_prediction \\
google/code\_x\_glue\_ct\_code\_to\_text \\
google-research-datasets/mbpp
\end{tabular} \\
\addlinespace
Other & 
\begin{tabular}[t]{@{}l@{}}
bigscience/P3 \\
wiki\_qa \\
Anthropic/persuasion \\
huggingface/cnn\_dailymail \\
allenai/qasper \\
openai/summarize\_from\_feedback \\
Salesforce/wikitext \\
Anthropic/llm\_global\_opinions \\
google-research-datasets/wiki\_split \\
google-research-datasets/aquamuse
\end{tabular} \\
\bottomrule
\end{tabular}
\caption{Datasets used for training router. Full citations can be found in Appendix A.}
\label{tab:comprehensive_training_datasets}
\end{table}

\begin{table}
\centering
\begin{tabular}{lr}
\toprule
\textbf{Domain} & \textbf{Number of Entries} \\
\midrule
Health & 100,000 \\
Math   & 113,611 \\
Science   & 224,885 \\
Coding & 572,636 \\
Other  & 700,000 \\

\bottomrule
\end{tabular}
\caption{Final data distribution across domains from datasets}
\label{tab:final_data_distribution}
\end{table}

\subsection{Experts}

\subsubsection{Expert Selection}

Our research use a combination of domain-specific and general-purpose models to create a system of expert agents. The selection of these models was primarily based on the availability of high-quality, open-source options that demonstrated superior performance in their respective domains. We utilized two sets of models: a ``medium'' set with larger parameter counts, and a ``small'' set with more compact models.

\subsection*{\textbf{Medium Model Set (\( \leq \)73B parameters)}}
The following models were chosen as the experts for our medium model:

\begin{itemize}
    
    \item \textbf{Health}: Palmyra-health-70B~\citep{team_introducing_2024}
    \item \textbf{Math}: Qwen2.5-72B-Math-Instruct~\citep{yang_qwen2_2024}
    \item \textbf{Science}: Qwen2.5-72B-Instruct~\citep{yang_qwen2_2024}
    \item \textbf{Coding}: Qwen2.5-72B-Instruct~\citep{yang_qwen2_2024}
    \item \textbf{Other}: Meta-Llama-3.1-70B-Instruct~\citep{dubey_Llama_2024}
\end{itemize}

\subsubsection*{Small MoDEM Model Set (\( \leq \)8B parameters)}

We also explored a set of smaller models, each with less than 8B parameters:

\begin{itemize}
    \item \textbf{Health}: Meta-Llama-3.1-8B-Instruct~\citep{dubey_Llama_2024}
    \item \textbf{Math}: Qwen2.5-Math-7B-Instruct~\citep{yang_qwen2_2024}
    \item \textbf{Science}: Qwen2.5-7B-Instruct~\citep{yang_qwen2_2024}
    \item \textbf{Coding}: Qwen2.5-Coder-7B~\citep{hui_qwen25-coder_2024}
    \item \textbf{Other}: Meta-Llama-3.1-8B-Instruct~\citep{dubey_Llama_2024}
\end{itemize}

The selection of models was based on evaluating across different domains, where we chose the best-performing models for each domain. In almost all cases, we found that modern models specialized in a specific domain significantly outperformed general-purpose models of the same size \cite{yang_qwen2_2024}. For instance, the Palmyra models excelled in health \cite{team_introducing_2024}, while the Qwen2.5-Math model proved to be the most effective for mathematical tasks \cite{yang_qwen2_2024}.

In cases where domain-specific models were not available, we defaulted to strong general-purpose models to maintain consistency across the system. Models like Meta-Llama-3.1 served as reliable baselines, ensuring good performance even in the absence of specialized options.

\subsection{Prompting}

We use zero-shot prompting with chain of thought \cite{wei_chain--thought_2023} to prompt each expert to answer questions in the benchmarks (Section \ref{sec:benchmarks}).\footnote{We use the following prompt: \textit{Solve the following problem step by step, explaining each step clearly to ensure the reasoning process is well-justified}. For multiple-choice questions, we have an additional sentence appended to the previous prompt: \textit{Clearly state which multiple choice option you pick}.} Full prompts can be found in appendix B

\begin{table}
\centering
\begin{tabular}{lr}
\toprule
\textbf{Category} & \textbf{Accuracy} \\
\midrule

Health   & 81.18\% \\
Math   & 96.63\% \\
Science   & 83.02\% \\
Coding & 77.42\% \\
Other  & 52.94\% \\

\midrule
\textbf{Overall} & 81.00\% \\
\bottomrule
\end{tabular}
\caption{Router Classification Results on MMLU.}
\label{tab:mmlu_results_percentages}
\end{table}

\section{Results}

\begin{table*}[t]\small
\centering

\begin{tabular}{l@{\hspace{1cm}}l@{\hspace{1cm}}l@{\hspace{1cm}}l@{\hspace{1cm}}l}
\toprule
\textbf{Domain} & \textbf{Benchmark} & \textbf{Llama 3.1 70B} & \textbf{Medium (<73B)} & \textbf{Improvement} \\
\midrule
Multi-domain    & MMLU              & 86.0\%   & \textbf{87.7\%}  & +1.7\%  \\
    & MMLU Pro          & 58.0\%   & \textbf{63.4\%}  & +5.4\%  \\
Coding          & HumanEval         & 80.5\%*   & \textbf{86.5\%*}  & +6.0\%  \\
Science          & GPQA         & 46.1\%   & \textbf{48.4\%}  & +2.3\%  \\
Math            & College Math      & 42.5\%*   & \textbf{49.5\%*}  & +7.0\%  \\
            & MATH              & 65.7\%*   & \textbf{85.9\%*}  & +20.2\% \\
            & GSM8k             & 94.1\%*   & \textbf{95.9\%*}  & +1.8\%  \\
            & Olympiad Bench    & 27.7\%*   & \textbf{49.0\%*}    & +21.3\% \\
\bottomrule
\end{tabular}
\caption{Comparison of Llama 3.1 70B vs.\ medium MoDEM (\( \leq \)73B) on various benchmarks. An asterisk (*) indicates numbers sourced from another paper. See Section 4.2 for further explanation.}
\label{tab:medium_modem_results}
\end{table*}

\begin{table*}[t]
\centering

\begin{tabular}{l@{\hspace{1cm}}l@{\hspace{1cm}}l@{\hspace{1cm}}l@{\hspace{1cm}}l}
\toprule
\textbf{Domain} & \textbf{Benchmark} & \textbf{Llama 8B} & \textbf{Small (<8B)} & \textbf{Improvement} \\
\midrule
Multi-domain    & MMLU              & 73.0\%   & \textbf{76.2\%}  & +3.2\%  \\
    & MMLU Pro          & 40.4\%   & \textbf{46.5\%}  & +6.1\%  \\
Coding          & HumanEval         & 72.6\%*   & \textbf{88.4\%*}  & +15.8\% \\
Science          & GPQA         & 32.6\%   & \textbf{35.0\%}  & +2.4\%  \\
Math            & College Math      & 33.8\%*   & \textbf{46.8\%*}  & +13.0\% \\
           & MATH              & 47.2\%*   & \textbf{83.6\%*}  & +36.4\% \\
            & GSM8k             & 76.6\%*   & \textbf{95.2\%*}  & +18.6\% \\
            & Olympiad Bench    & 15.4\%*   & \textbf{41.6\%*}  & +26.2\% \\
\bottomrule
\end{tabular}
\caption{Comparison of Llama 8B vs.\ small MoDEM (\( \leq \)8B) on various benchmarks. An asterisk (*) indicates numbers sourced from another paper. See Section 4.2 for further explanation.}
\label{tab:small_modem_results}
\end{table*}

\subsection{Router Performance}

We evaluated our router on the test set of the datasets used for training, and it achieved an average accuracy of 97\%, illustrating its high reliability in routing queries for tasks similar to those it was fine-tuned on. We next assessed the router's performance on the MMLU to test its ability to generalize to out-of-distribution data. We manually mapped the MMLU domains into our chosen domains.\footnote{Recall that MMLU was not used in the training data for the router.} Table \ref{tab:mmlu_results_percentages} presents the results. We generally see strong performance for the specialised domains, although for ``Other'' the performance is a little lower. The latter observation is perhaps not too surprising, it's a ``catch all'' domain that doesn't have a concrete definition and so it's difficult to have training data that captures the full data distribution. Overall these results suggest that the router generalises well and is sufficiently reliable as a domain router.

We manually assessed some of the error cases and found that some mis-classifications are due to domain-ambiguity. To give an example:
\begin{itemize}
    \item \textbf{Example} "A burial site where the body is allowed to decompose naturally without a casket is called a \_\_\_\_ cemetery." \\
    \textbf{True Domain:} Health, \textbf{Predicted:} Other

\end{itemize}

\subsection{MoDEM Performance}

We present the full results in Table  \ref{tab:medium_modem_results} and \ref{tab:small_modem_results} for medium and small MoDEM respectively.
For baseline comparisons, we used the Llama 3.1 instruct models, which are generally considered SoTA for open source models. In instances where the same prompting techniques (zero-shot with Chain of Thought) were employed, we use reported outcomes (denoted by an asterisk in the tables) due to computational limitations and challenges associated with evaluating certain benchmarks (e.g.\ the test set is not open-source).\footnote{For these benchmarks, we found in practice over 98\% of the prompts were routed to a single model (e.g.\ 98.4\% of Math benchmark was routed to our math expert) and so the results would be reasonably close to those we would obtain if we ran them ourselves.}

Concretely, we ran the MMLU, MMLU-Pro and GPQA benchmark results ourselves for the baseline. But for all other benchmarks (HumanEval, College Math, Math, GSM8k and Olympiad Bench)

we sourced the results from the Qwen-2.5 Technical Report \cite{yang_qwen2_2024} and the Llama 3.1 Technical Report \cite{dubey_Llama_2024}.

\begin{table*}
\centering
\begin{tabular}{lccc}
\toprule
\textbf{Model} & \textbf{MMLU Accuracy (\%)} & \textbf{Parameters} & \textbf{Input Tokens (\$/million tokens)} \\
\midrule
Llama 3.1 405B     & 88.6  & 405B   & 5.00 \\
\textbf{Medium MoDEM}    & \textbf{87.7}  & \textbf{<73B}   & \textbf{0.92} \\
Qwen 2.5-72B     & 86.1  & 72B    & 0.9 \\
Llama 3.1 70B      & 86.0  & 70B    & 0.88 \\
Mixtral-8x22B  & 77.5  & 8x22B   & 1.20 \\
\bottomrule
\end{tabular}
\caption{Comparison of medium MoDEM vs.\ leading models in terms of estimated inference cost.}
\label{tab:medium_modem_comparison} 

\end{table*}

\begin{table*}
\centering
\begin{tabular}{lccc}
\toprule
\textbf{Model} & \textbf{MMLU Accuracy (\%)} & \textbf{Parameters} & \textbf{Input Tokens (\$/million tokens)} \\
\midrule
Llama 3.1 70B      & 86.0  & 70B    & 0.88 \\
\textbf{Small MoDEM}           & \textbf{76.2}  & \textbf{<8B}    & \textbf{0.22} \\
Llama 3.1 8B         & 73.0  & 8B     & 0.18 \\
Mixtral-8x7B Instruct & 70.6  & 8x7B   & 0.60 \\
Gemma2-9B            & 69.2  & 9B     & 0.30 \\
Mistral-7B           & 62.5  & 7B     & 0.20 \\
\bottomrule
\end{tabular}
\caption{Comparison of small MoDEM vs.\ leading models in terms of estimated inference cost.}
\label{tab:small_modem_comparison} 

\end{table*}

MoDEM demonstrate consistent performance gain across all evaluated benchmarks when compared to their respective baselines. This consistent improvement highlights the effectiveness of our domain-specialized models and the strength of the routing system in accurately selecting the appropriate expert for each task.
For the math domain in particular, MoDEM delivered substantial improvements. The performance gains in these areas show the clear advantage of domain-specific training and highlight the effectiveness of our approach to model specialization.
In tasks involving multi-domain knowledge and reasoning (MMLU and MMLU-Pro), both small and medium MoDEM still show improvement over the baseline, demonstrating MoDEM is versatile across different domains.

\subsection{Cost and Efficiency Analysis }
To evaluate the efficiency of our model, we compared its performance and inference costs with other leading models. All costs are based on Together AI \cite{noauthor_notitle_nodate} figures where possible. For models not publicly hosted we based price off models of similar size. At the time of publishing the Qwen 2.5 models were not publicly hosted so we defaulted to the Qwen 2 prices. Palmyra-Health was also not hosted on TogetherAi so we use the price of the Writer API.
For our router cost we assumed pricing based off other Bert based models of similar size being hosted. We assumed \$0.03 per million tokens for the router cost. The reported cost for our models were based off the average over the MMLU dataset. Prices may vary slightly depending on dataset due to different experts models having different inference costs.

MMLU results are in Table \ref{tab:medium_modem_comparison} and \ref{tab:small_modem_comparison} for medium and small MoDEM  respectively.
Our models demonstrate a superior price-to-performance ratio compared to the leading models. Both  medium and small MoDEM deliver higher accuracies across benchmarks while maintaining competitive or lower inference costs, showcasing significant improvements in cost-effectiveness.

For small MoDEM in particular, we see that it has a much better performance compared to similar sized models. For medium MoDEM, its performance is close to a much larger model (Llama 405B), even though it is 5-6 times smaller and cheaper. Together these results illustrate the scalability and effectiveness of our approach across a range of model sizes.

\section{Discussion}
The results of our study on mixture of experts with domain-specific routing suggest a potential shift in the development and deployment of large language models (LLMs). This section explores the implications of our findings, their broader impact on the field of artificial intelligence, and potential directions for future research.

\subsection{Potential Shift in Model Development}

Our research demonstrates that combining domain routing with models fine-tuned for specific domains can significantly outperform base models of the same size, leading to a more favorable performance-to-cost ratio. This challenges the current trend of developing increasingly large, general-purpose models and instead points towards a future where AI systems consist of an ecosystem of smaller, highly specialized models coupled with intelligent routing mechanisms.

This shift parallels how human expertise is organized in society, where specialists in various fields collaborate to solve complex problems. In the context of AI, this approach could result in:

\begin{itemize}
    \item More efficient resource utilization
    \item Reduced computational costs
    \item Superior performance in domain-specific tasks
    \item Increased interpretability and control over model outputs
\end{itemize}

As compute bottlenecks continue to constrain the development of ever-larger models, the transition towards domain-specific models may become necessary to sustain progress in LLM capabilities and performance. By optimizing resources and leveraging domain expertise, this approach holds promise for maintaining the current rate of advancements in the field.

Our approach holds significant potential for future improvement. As the AI community develops more specialized, high-performance models, we anticipate substantial increases in the overall capabilities of our system. The current performance represents a lower bound of what’s achievable, and as specialized models trained on domain-specific data emerge, it will benefit our mixture of experts routing approach.

We want to also highlight that MoDEM's domain set is adaptable. As new specialized models in fields like legal or environmental science become available, they can be easily integrated by updating the router and adding relevant expert models. Existing domains can also be refined or consolidated based on performance analysis, ensuring continued efficiency. Additionally, hierarchical domain structures, such as broad categories with more specific sub-domains, could further enhance routing precision.
This adaptable approach ensures our system evolves with AI developments, providing a scalable framework for continuous improvement aligned with real-world needs.

\subsection{Implications for AI Deployment}

Our findings reveal that domain-specific models with fewer parameters can match or outperform larger general-purpose models like Llama 405B, carrying important implications for AI deployment. This approach delivers state-of-the-art performance at a fraction of the inference cost, drastically reducing computational overhead while maintaining high-quality results. It opens opportunities for cost-effective AI deployment, particularly in resource-constrained settings where large models are impractical.

\subsection{Future Research Directions}
Our findings highlight several promising research directions using mixture of experts. Key challenges include developing better routing techniques, such as improving domain selection accuracy and scaling to more domains. Expanding domain-specific models to cover a wider range of tasks will also increase the system's applicability across industries. Cross-domain integration and dynamic model selection could enhance handling of complex queries by combining outputs from multiple experts in real time. Additionally, introducing difficulty-based routing within each domain could optimize resource use, directing simpler queries to smaller models and complex ones to larger models, improving cost-effectiveness and performance.

\section{Conclusion}
This study demonstrates the effectiveness of combining domain-specific expert models with routing to enhance the performance and efficiency of large language models. Our approach consistently outperformed baseline models across various benchmarks, with strong improvement in specialized domains such as mathematics. Both our small and medium MoDEM  achieved superior performance-to-cost ratios compared to larger, general-purpose models, highlighting the potential for significant efficiency gains in AI deployment.

This research demonstrates a promising new direction in the field of artificial intelligence: the combination of domain-specific models with intelligent routing systems. The study’s findings suggest that this approach can lead to significant improvements in both performance and cost-efficiency compared to traditional large language models. These findings point to a potential shift in AI development and deployment. Rather than focusing solely on creating increasingly large general-purpose models, the future may lie in developing ecosystems of smaller, highly specialized models coupled with sophisticated routing systems. This approach could lead to more efficient resource utilization, reduced computational costs, and superior performance in domain-specific tasks.

\section*{Limitations}

It's important to note that our selection was constrained by the current landscape of available open-source, domain-specific models. The field of AI is rapidly evolving, and the development of specialized models is a relatively recent trend. As such, our study represents an initial exploration into the potential of combining domain experts with intelligent routing.

Additionally, we were somewhat limited by the lack of public APIs for certain models, making it challenging to run direct benchmarks. This constraint forced us to rely on benchmarks reported in other studies, which may not have fully captured the performance nuances in our specific use case. As more models become accessible and standardized benchmarking tools evolve, future iterations of our research will likely benefit from more comprehensive and direct performance evaluations.

\appendix

\section*{Appendix A: Dataset Citations}
\addcontentsline{toc}{section}{Appendix C: Dataset Citations}

Below is a list of citations for the datasets used in our study, organized by domain:

\begin{itemize}
    \item \textbf{Math}
    \begin{itemize}
        \item TIGER-Lab/MathInstruct: \cite{yue2023mammoth}
        \item lighteval/MATH: \cite{yue_mammoth_2023}
        \item allenai/math\_qa: \cite{amini-etal-2019-mathqa}
        \item openai/gsm8k: \cite{cobbe2021gsm8k}
        \item camel-ai/math: \cite{li2023camel}
        \item meta-math/MetaMathQA: \cite{yu2023metamath}
        \item deepmind/math\_dataset/algebra\_\_linear\_1d: \cite{2019arXiv}
        \item deepmind/math\_dataset/algebra\_\_polynomial\_roots: \cite{2019arXiv}
        \item deepmind/aqua\_rat: \cite{ling2017program}
        \item AI4Math/MathVerse: \cite{zhang2024mathverse}
    \end{itemize}
    \item \textbf{Health}
    \begin{itemize}
        \item nlpaueb/biomrc: \cite{pappas-etal-2020-biomrc}
        \item iari/HumGen\_Clinical\_Notes: augmented-clinical notes
        \item medmcqa: \cite{pmlr-v174-pal22a}
        \item lavita/ChatDoctor-HealthCareMagic-100k: \url{https://huggingface.co/datasets/lavita/ChatDoctor-HealthCareMagic-100k}
    \end{itemize}
    \item \textbf{Science}
    \begin{itemize}
        \item bigbio/pubmed\_qa: \cite{jin2019pubmedqa}
        \item derek-thomas/ScienceQA: \cite{lu2022learn}
        \item allenai/sciq: \cite{SciQ}
        \item bigscience/P3: \cite{sanh2021multitask}
        \item ai2\_arc: \cite{allenai:arc}
        \item nlpaueb/biomrc: \cite{pappas-etal-2020-biomrc}
        \item allenai/scitldr: \cite{cachola2020tldr}
        \item tdiggelm/climate\_fever: \cite{diggelmann2020climatefever}
        \item medmcqa: \cite{pmlr-v174-pal22a}
        \item Idavidrein/gpqa: \cite{rein_gpqa_2023}
        \item allenai/scifact: \cite{wadden-etal-2020-fact}
        \item allenai/scirepeval: \cite{wadden-etal-2020-fact}
    \end{itemize}
    \item \textbf{Coding}
    \begin{itemize}
        \item codeparrot/apps: \cite{hendrycksapps2021}
        \item bigcode/the-stack: \cite{Kocetkov2022TheStack}
        \item nuprl/MultiPL-E: \cite{cassano:multipl-t}
        \item code\_x\_glue\_ct\_code\_to\_text: \cite{husain2019codesearchnet}
        \item deepmind/code\_contests: \cite{li2022competition}
        \item huggingface/codecompetitions: \cite{li2022competition}
        \item openai/openai\_humaneval: \cite{chen2021evaluating}
        \item bigcode/humanevalpack: \cite{muennighoff2023octopack}
        \item defect\_prediction: \cite{zhou2019devign}
        \item google/code\_x\_glue\_ct\_code\_to\_text: \cite{husain2019codesearchnet}
        \item google-research-datasets/mbpp: \cite{austin2021program}
    \end{itemize}
    \item \textbf{Other}
    \begin{itemize}
        \item bigscience/P3: \cite{sanh2021multitask}
        \item wiki\_qa: \cite{yang-etal-2015-wikiqa}
        \item Anthropic/persuasion: \cite{durmus2024persuasion}
        \item huggingface/cnn\_dailymail: \cite{see-etal-2017-get}
        \item allenai/qasper: \cite{Dasigi2021ADO}
        \item openai/summarize\_from\_feedback: \cite{stienon2020learning}
        \item Salesforce/wikitext: \cite{merity2016pointer}
        \item Anthropic/llm\_global\_opinions: \cite{durmus2023measuring}
        \item google-research-datasets/wiki\_split: \cite{botha-etal-2018-learning}
        \item google-research-datasets/aquamuse: \cite{kulkarni2020aquamuse}
    \end{itemize}
\end{itemize}

\section*{Appendix B: Prompting Techniques}
\addcontentsline{toc}{section}{Appendix B: Prompting Techniques}

\subsection*{For Prompting the Model}

\textbf{Prompt:} \\
\textit{Solve the following problem step by step, explaining each step clearly to ensure the reasoning process is well-justified. Clearly state which multiple choice option you pick.} \\

Input:
\begin{verbatim}
{question}
\end{verbatim}

\vspace{1em} 

\subsubsection*{For Our LLM Evaluation}

Prompt:
\textit{You will be given a ground truth answer and a model answer. Please output ACCURATE if the model answer matches the ground truth answer or INACCURATE otherwise. Please only return ACCURATE or INACCURATE. It is very important for my job that you do this.} \\

Input Format:
\begin{verbatim}
<GroundTruthAnswer>
{correctAnswer}
</GroundTruthAnswer>

<ModelAnswer>
{predictedAnswer}
</ModelAnswer>
\end{verbatim}

\end{document}